\documentclass{CUP-JNL-DTM}%
\usepackage{graphicx}
\usepackage{multicol,multirow}
\usepackage{amsmath,amssymb,amsfonts}
\usepackage{mathrsfs}
\usepackage{amsthm}
\usepackage{rotating}
\usepackage{appendix}
\usepackage[natbib,style=apa]{biblatex}
\usepackage{ifpdf}
\usepackage[T1]{fontenc}
\usepackage{newtxtext}
\usepackage{newtxmath}
\usepackage{textcomp}
\usepackage{xcolor}
\usepackage{lipsum}
\usepackage[colorlinks,allcolors=blue]{hyperref}

\usepackage{algorithm}
\usepackage{algpseudocode}
\usepackage{subcaption}
\usepackage{lmodern}

\DeclareCaptionLabelFormat{bold}{\textbf{(#2)}}
\theoremstyle{definition}

\numberwithin{equation}{section}

\jname{}
\articletype{Application Paper}
\jyear{2023}

\begin{document}

\begin{Frontmatter}

\title[Article Title]{Low-Cost Tree Crown Dieback Estimation Using Deep Learning-Based Segmentation}

\author[1]{Allen, M.J.$^{*,}$}
\author[2,3]{Moreno-Fernández, D.}
\author[2,4]{Ruiz-Benito, P.}
\author[5,6]{Grieve, S.W.D.}
\author[1]{Lines, E.R.}

\authormark{Author Name1 \textit{et al}.}

\address[1]{\orgdiv{}\orgname{Department of Geography, University of Cambridge}, \orgaddress{\city{Cambridge}, \postcode{CB2 3EN}, \state{} \country{UK}}}

\address[2]{\orgdiv{Forest Ecology and Restoration Group}, \orgname{Departamento de Ciencias de la Vida, Universidad de Alcal\'a}, \orgaddress{\city{Madrid}, \postcode{28805}, \state{}  \country{Spain}}}

\address[3]{\orgdiv{}\orgname{Institute of Forest Sciences (INIA-CSIC), Crta. de la Coruña km 7.5}, \orgaddress{\city{Madrid}, \postcode{28040}, \state{}  \country{Spain}}}

\address[4]{\orgdiv{Environmental Remote Sensing Research Group}, \orgname{Departamento de Geolog\'ia, Universidad de Alcal\'a}, \orgaddress{\city{Alcal\'a de Henares}, \postcode{28801}, \state{}  \country{Spain}}}

\address[5]{\orgdiv{}\orgname{School of Geography, Queen Mary University of London}, \orgaddress{\city{London}, \postcode{E1 4NS}, \state{} \country{UK}}}

\address[6]{\orgdiv{}\orgname{Digital Environment Research Institute, Queen Mary University of London}, \orgaddress{\city{London}, \postcode{E1 1HH}, \state{} \country{UK}}}

\address[*]{\orgdiv{}\orgname{Corresponding author}, \email{mja78@cam.ac.uk}}

\authormark{Allen et al.}


\abstract{
    The global increase in observed forest dieback, characterised by the death of tree foliage, heralds widespread decline in forest ecosystems. This degradation causes significant changes to ecosystem services and functions, including habitat provision and carbon sequestration, which can be difficult to detect using traditional monitoring techniques, highlighting the need for large-scale and high-frequency monitoring. Contemporary developments in the instruments and methods to gather and process data at large-scales mean this monitoring is now possible. In particular, the advancement of low-cost drone technology and deep learning on consumer-level hardware provide new opportunities. Here, we use an approach based on deep learning and vegetation indices to assess crown dieback from RGB aerial data without the need for expensive instrumentation such as LiDAR. We use an iterative approach to match crown footprints predicted by deep learning with field-based inventory data from a Mediterranean ecosystem exhibiting drought-induced dieback, and compare expert field-based crown dieback estimation with vegetation index-based estimates. We obtain high overall segmentation accuracy (mAP: 0.519) without the need for additional technical development of the underlying Mask R-CNN model, underscoring the potential of these approaches for non-expert use and proving their applicability to real-world conservation. We also find colour-coordinate based estimates of dieback correlate well with expert field-based estimation. Substituting ground truth for Mask R-CNN model predictions showed negligible impact on dieback estimates, indicating robustness. Our findings demonstrate the potential of automated data collection and processing, including the application of deep learning, to improve the coverage, speed and cost of forest dieback monitoring.
}

\end{Frontmatter}

\section*{Impact Statement}
The global increase in forest dieback threatens vital ecosystem services, including habitat provision and carbon sequestration. Traditional field and drone-based monitoring techniques are not cost-effective at large scales. We leverage advancements in low-cost drones and deep learning to assess crown dieback from drone-captured RGB aerial data, eliminating the need for costly tools like LiDAR and laborious fieldwork. We achieve high accuracy matching predicted crown footprints with actual forest inventory, demonstrating the effectiveness of these methods for non-expert use in conservation. Our vegetation index-based dieback estimates aligned well with expert field-based estimates, validating this approach. This work enhances forest monitoring by improving coverage, accuracy and speed, and reducing costs, thereby providing a promising tool to address large-scale conservation challenges.

\section{Introduction}\label{sec:intro}
Forest dieback, the unseasonal loss of crown foliage or mortality of many trees \mbox{\citep{mueller-dombois_forest_1988}}, is an early indicator of declining health in forest ecosystems. This degradation directly affects ecosystem services and functions such as carbon sequestration \citep{baccini_tropical_2017} and habitat provision \citep{watson_exceptional_2018}. Various factors, including the spread of pests, pathogens, and the intensification of drought conditions due to climate change, contribute to the increasing prevalence of crown dieback around the world (\cite{carnicer_widespread_2011}; \cite{liebhold_highly_2013}; \cite{senf_excess_2020}). These changes in forest health and structure can result in ecologically significant shifts in composition and function over relatively short time periods \citep{allen_underestimation_2015}, which can be difficult to detect using traditional sampling strategies \citep{mcmahon_importance_2019}. 
\\\\
The global increase in dieback and its potential impacts on ecosystem services and functions highlight the urgent need for large-scale, high-frequency monitoring. Large-scale, high-frequency monitoring of forest health and structure would enhance our ability to detect issues such as pest outbreaks, disease, or stress caused by climate change \citep{mcmahon_importance_2019} at early stages - which would, in turn, support better-informed decision-making and policy development related to forest management, climate change mitigation, and biodiversity conservation \citep{asner_spectranomics_2016}. Moreover, large-scale monitoring techniques enable practitioners to better understand the factors driving dieback, informing more effective and targeted management strategies, increasing resilience to these threats before they emerge. The instruments and methods to gather and process data for large-scale, rapid monitoring at the level of individual trees have, however, only been recently developed \mbox{(\cite{diez_deep_2021}; \cite{kattenborn_review_2021}; \cite{ecke_uav-based_2022})}. %
\\\\
Remote sensing can provide high resolution spatiotemporal data coverage to meet these needs. Leaf reflectance, including within the visible spectrum, is sensitive to changes in plant physiology, chemistry and structure due to insect and pathogen infestation as well as water stress \mbox{\citep{huang_remote_2019}}. Such changes in leaf reflectance are observable in the canopy from remotely sensed data \mbox{\citep{huang_remote_2019}} so high-resolution satellite imagery can likely detect  defoliation missed by field-based surveys - for example, in low-severity areas \citep{bright_using_2020}. Many existing approaches, however, rely on large spatial averages of stress indicators - at the approximately 10m spatial scale of common non-commercial satellite data products such as Sentinel 1/2 \citep{lastovicka_sentinel-2_2020} or Landsat \citep{zhang_tracking_2021} - and therefore cannot capture information on individual tree health. Tree level information is crucial to understand future forest dynamics - particularly where individual trees may exhibit significantly different responses, such as for disease (\cite{fensham_unprecedented_2021}; \mbox{\cite{hurel_genetic_2021}}; \cite{kannaste_impacts_2023}) or drought (\cite{teskey_responses_2014}; \cite{chen_strategies_2022}; \mbox{\cite{fernandez-de-una_role_2023}}). New approaches, leveraging the increasing availability of high resolution data ($\leq0.3$m), are required to identify individual trees within imagery for monitoring.
\\\\
Traditional segmentation methods - based on handcrafted algorithms and manual feature engineering \mbox{\citep{baatz_multiresolution_2000}} - have been used to extract individual tree crown (ITC) polygons from aerial imagery for use in some downstream tasks useful for monitoring. For example, \citet{onishi_explainable_2021} applied unsupervised multiresolution segmentation \citep{baatz_multiresolution_2000} to RGB (Red, Green, Blue) data of a dense forest canopy in urban Japan, combined with digital surface models (DSMs) to segment ITCs, and classified the results by species or functional type. Although the automatically extracted ITCs contained mostly canopy of the same species, the segmentation algorithm fragmented or merged $76\%$ of tree crowns with neighbouring individuals within that species. This method did not require manual labelling, but the low intra-specific segmentation accuracy means it is not comparable to traditional inventory-based monitoring, as the ITC polygons cannot be used to accurately extract measurements relating to individual trees - such as measurement of stress responses including dieback. Since these responses vary substantially, even within species (\cite{hurel_genetic_2021}; \cite{fernandez-de-una_role_2023}), accurate segmentation is crucial. We suggest that a deep learning-based approach is likely to yield better results on data with structurally complex canopies from natural forests, where crowns intersect substantially both within and across species. 
\\\\
Many approaches applying deep learning to segment individual tree crowns exist \mbox{\citep{diez_deep_2021}}. Deep learning was applied to perform segmentation through bounding box delineation (typically referred to as object detection  in machine learning literature - for example, in \cite{ren_faster_2016}) of trees in both open and closed forest canopies from RGB images in \citet{weinstein_individual_2019}. Object detection was performed with a single-stage deep learning detector (\cite{he_deep_2015}; \cite{lin_focal_2018}), achieving precision of 0.69 and recall of 0.61 in forests from California, at a National Ecological Observation Network (NEON) site. The network was pre-trained using a very large number of noisy labels extracted from LiDAR data and fine tuned on a smaller number of manual annotations. Precise delineation of individual crowns (typically referred to as instance segmentation in machine learning literature - for example, in \cite{he_mask_2017}) through deep learning has also been applied produce the polygons required to quantify stress accurately at the individual level \mbox{(\cite{chiang_deep_2020}}; \mbox{\cite{hao_automated_2021}}; \mbox{\cite{yang_detecting_2022}}; \mbox{\cite{sandric_trees_2022}}; \mbox{\cite{sani-mohammed_instance_2022}}; \mbox{\cite{ball_accurate_2023}}). Throughout this work, we refer to this type of segmentation, when applied to tree crowns, as `ITC delineation' to provide clarity to practitioners unfamiliar with machine learning literature, although `instance segmentation' could be used interchangeably. \citet{hao_automated_2021}, for example, used Mask R-CNN to perform ITC delineation of tree crowns from multispectral 2D imagery coupled with a photogrammetry-derived canopy height model (CHM) in a Chinese fir (\textit{Cunninghamia lanceolata}) plantation, and achieved an F1 score of 0.85. It was demonstrated that these delineations could be used successfully for a simple downstream task - the extraction of individual tree heights by superposing segmentation on the CHM data. The use of data from plantations, however, is unlikely to yield results that are representative of those from natural woodland; stems are typically evenly aged and spaced, and do not have structurally complex canopy. \mbox{\citet{sandric_trees_2022}} similarly segmented trees with Mask R-CNN in partly artificial contexts - across 5 different species in temperate and Mediterranean orchards. \citet{yang_detecting_2022} demonstrated that ITC delineation is possible in canopy of greater structural complexity, showing some crown intersection and overlap, in the heavily managed environment of Central Park in New York City, USA - although delineation was not verified using  ground data - and further demonstrated that a simple structural measurement (crown area) could be replicated compared to manual measurement from aerial data by using these delineations downstream. \citet{ball_accurate_2023} further demonstrated accurate ITC delineation in tropical forests with a structurally complex canopy, using Mask R-CNN, although differences in leaf reflectance and canopy structure corresponding to species differences may facilitate segmentation more easily than in monospecific canopy, where spectral variation may be relatively lower. \mbox{\citet{sani-mohammed_instance_2022}} were similarly able to perform ITC delineation using Mask R-CNN in natural temperate forest in Bavaria although only segmented dead trees, which may be easier to identify due to lack of foliage. The use of automatically extracted crown footprints for complex applications such as forest health monitoring, however, is less comprehensively explored.
\\\\
Several studies exploiting deep learning for automated forest health measurement have emerged in recent years. Some work explores the use of classical computer vision, based on manually engineered algorithms, to perform ITC delineation and then applies deep learning to classify damage levels. \mbox{\citet{safonova_detection_2019}}, for example, applied manual filtering and thresholding to extract rectangular patches of treetops and applied a number of CNN-based classifiers downstream to classify damage into one of four levels. \citet{nguyen_individual_2021} similarly applied a manually engineered algorithm for patch extraction, instead based on normalised Digital Surface Models (nDSMs). The extracted patches were classified by damage categorically. Although classification of tree health into multiple damage levels is a promising approach, as dieback is a symptom with several continuous stages \citep{ciesla_decline_1994}, such approaches may be limited by their segmentation accuracy, particularly where individual trees differ significantly in their response, within and across species. Other works use deep learning-based segmentation for forest health assessment. \citet{schiefer_uav-based_2023} used a U-net to perform semantic segmentation (per-pixel classification of all pixels within a target image) of standing deadwood in UAV imagery and extrapolated this to the landscape level predictions on time series satellite imagery using Long Short Term Memory Networks (LSTMs). \citet{chiang_deep_2020} and \citet{sani-mohammed_instance_2022} both used Mask R-CNN to perform ITC delineation of dead tree crowns from aerial imagery. Where dieback is ongoing, however, it requires measurement on a continuous scale or by multiple categories, rather than binary classification \citep{ciesla_decline_1994}. \citet{sandric_trees_2022} used deep learning-based ITC delineation via Mask R-CNN to segment crown footprints, and performed post-hoc analysis using vegetation indices derived from colour space transformations to measure tree health on a continuous scale. The approach of using vegetation indices - which do not require additional human labelling - on footprints extracted via deep learning, shows great promise for dieback measurement, but measurements of crown health were not verified versus human measurement. Additionally, the use of plantation data may more easily facilitate this type of forest health measurement - although often monospecific, the canopy formed by artificially spaced trees is easily delineated, simplifying health monitoring. This structural simplicity may not reflect performance in many natural ecosystems.
\\\\
Here, we develop a new approach for early dieback detection from aerial RGB imagery, using deep-learning based ITC delineation and vegetation indices, and test it in an ecosystem experiencing dieback. Crucially, unlike in previous work \mbox{\citep{sandric_trees_2022}}, we verify our results by comparison to field-based dieback measurement by experts. We use drone data collected in a Mediterranean stone pine (\textit{Pinus pinea}) forest with a structurally complex monospecific canopy, where some individuals are showing signs of drought-induced crown dieback \citep{moreno-fernandez_interplay_2022}. The severity of drought is projected to increase across the Mediterranean region (\cite{dubrovsky_multi-gcm_2014}; \cite{hertig_regional_2017}), and poses a large scale threat to the functioning of this biodiversity hotspot. Protection through active monitoring and management of these ecosystems is required (\cite{fernandez-manjarres_forest_2018}; \cite{astigarraga_evidence_2020}) - but is currently limited by reliance on time-consuming expert manual inventories taken on the ground. In addition to offering a solution to conducting this monitoring at large scale and low cost, drone-based remote sensing is ideal for the monitoring of drought-induced crown dieback, as effects at crown extremities are more easily observable from above than  from traditional ground-based visual assessment. We develop a scalable approach based on deep learning and RGB vegetation indices to monitor this dieback at the individual level, and answer the following questions:

\begin{enumerate}
  \item Is ITC delineation possible in a structurally complex, monospecific canopy?
  \item Does individual health assessment based on vegetation indices \mbox{\citep{sandric_trees_2022} correlate with field-based estimation}, and to what degree is this assessment affected by the accuracy of deep learning-based ITC delineation?
\end{enumerate}

\section{Material \& Methodology}\label{sec:methods}
\subsection{Data}\label{sec:data}
\subsubsection{Study Area \& Inventory}
\begin{figure}[] 
\centering 
\begin{subfigure}[b]{0.49\textwidth}
    \includegraphics[width=\textwidth]{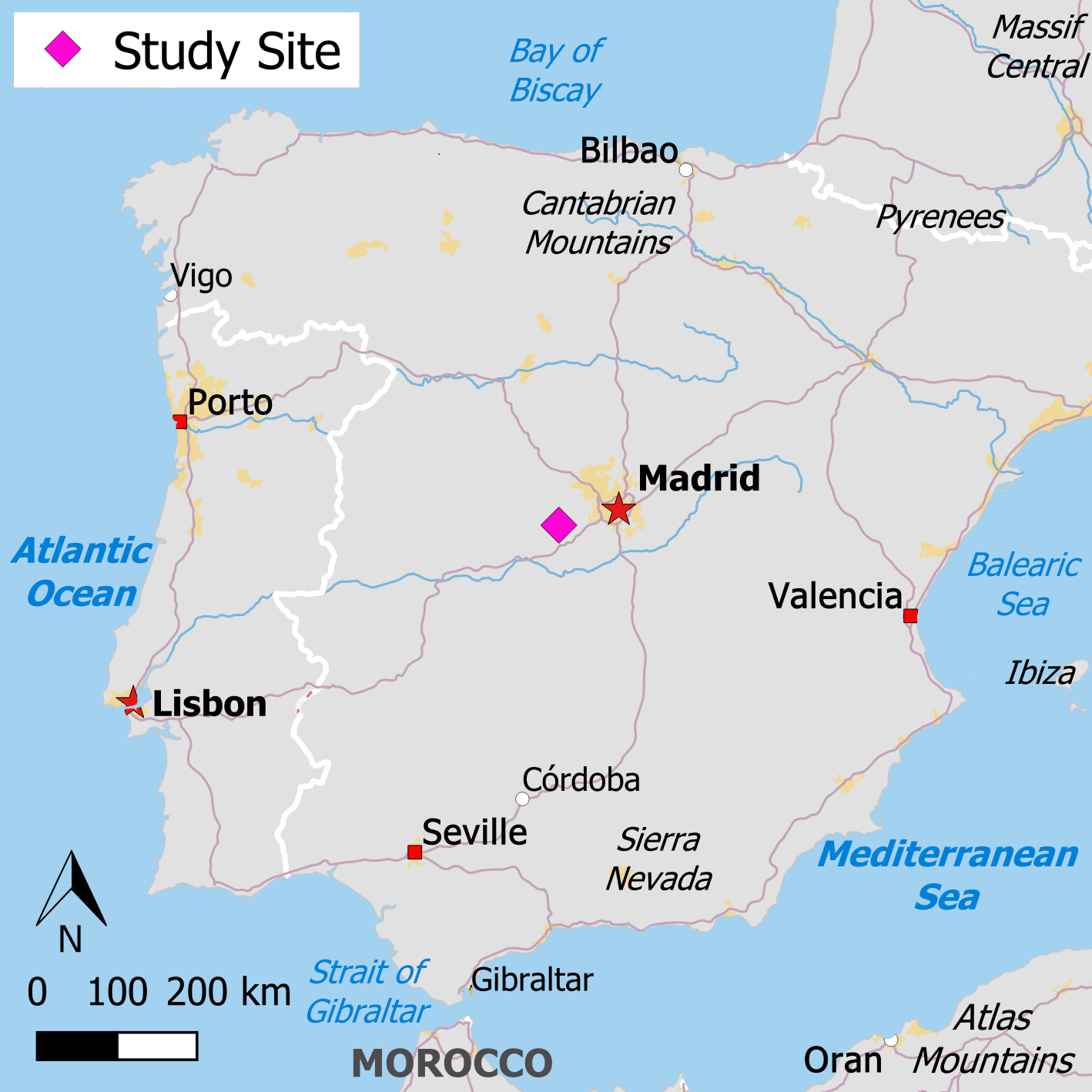}
    {\phantomsubcaption\label{fig:map-left}}
\end{subfigure}
\hfill
\begin{subfigure}[b]{0.49\textwidth}
    \includegraphics[width=\textwidth]{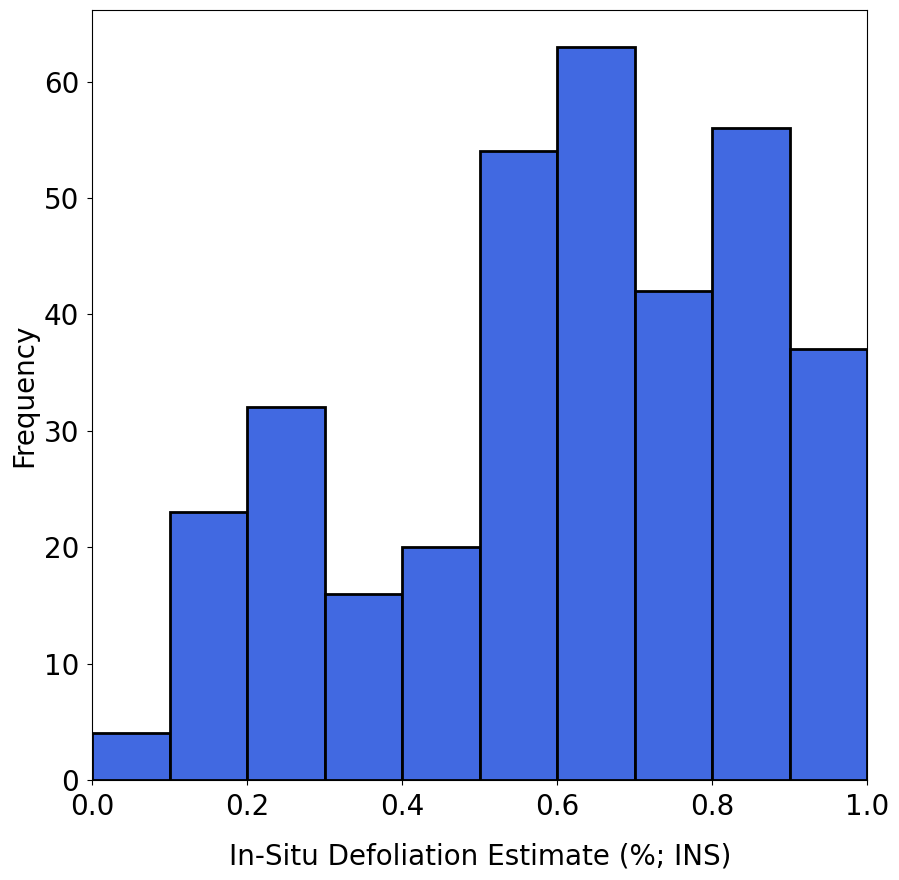}
    {\phantomsubcaption\label{fig:histogram}}
\end{subfigure}
\caption{\textbf{(a)} Map showing the location of the site described by our data in the Iberian peninsula. Map data from \citet{openstreetmap}. \textbf{(b)} Histogram of visually estimated defoliation (percentage needles missing) of adult trees in our field data. An estimate of 1.0 corresponds to a tree for which all foliage is dead, and 0.0 to a tree with completely healthy foliage. 453 adult trees (DBH > 7.5cm) were surveyed in total.}
\end{figure}

Our dataset covers 1500 ha total of \textit{Pinus pinea} forest, across nine areas, showing signs of climate-induced crown dieback. Drone and ground data were collected in Pinar de Almorox, Spain (40.27 \textdegree N, 4.36 \textdegree W) in May/June 2021. A map showing the location of the study site can be seen in Figure~\ref{fig:map-left}. The area is under a continental Mediterranean climate with a mean annual rainfall of 568 mm, a mean annual temperature of 14\textdegree C and altitude ranging from 500 to 850 m above sea level. In addition to the dominant canopy species, \textit{P. pinea}, a smaller number of \textit{Juniperus oxycedrus L.} and \textit{Quercus ilex} can be found in the midstory. The understory further contains \textit{Salvia rosmarinus L., Lavandula stoechas Lam.} and \textit{Lamiaceae} shrubs, as well as fallen deadwood. Nine distinct, non-overlapping areas were selected to show a gradient of defoliation, and within these inventory data was taken at 31 sample plots (three to four plots per area). Sample plots were circular and of 17 m radius, with a total of 453 adult (diameter at breast height, DBH, > 7.5 cm) trees surveyed in total. Each adult tree was assessed by experts in the field for dieback estimation (percentage defoliation). Crown dieback percentage was estimated visually (with 100\% corresponding to wholly defoliated), taking the average score from two experts for each tree within each plot. A histogram of dieback/defoliation percentage for all surveyed trees can be seen in Figure~\ref{fig:histogram}. Trunk locations within each plot were georeferenced using an Emlid Reach RS2 multi-band RTK Global Navigation Satellite System (GNSS) receiver \citep{ng_performance_2018}.  See \citet{moreno-fernandez_interplay_2022} for further analysis of dieback patterns at the site and additional data collected.
\\\\
Note that each area surveyed by the drone was not covered entirely by the plot data. Consequently, trees at the edges of each orthomosaic did not have in-situ defoliation estimates or geolocated trunk coordinates. These trees were still delineated manually from orthomosaic images at a later stage to provide training data for deep-learning based ITC delineation, but they were not considered in our analysis of dieback estimation.

\subsubsection{Drone Imagery}\label{sec:droneimagery}
Drone flights were used to obtain images over the nine non-overlapping areas (minimum area 16663m$^2$, maximum area 39736m$^2$), covering all plots, using a DJI Mavic Mini drone. The work was carried out with all relevant permissions, and all national drone regulations and location-specific regulations, including use of rotor guards, were observed. As this area is of high conservation value, particularly for birds of prey, the areas to survey were discussed and approved with land managers in advance of flights. All flights were performed at a fixed altitude 50m above take-off location, and carried out in May/June 2021 at the same time as ground data collection. Flight start times varied from 09:30 to 17:00 (local time) and conditions varied from overcast to clear, resulting in significant differences in lighting conditions and shadowing between drone flights. Raw images of size 4000$\times$2250px were taken from both nadir and oblique (55\textdegree$\,$below horizontal; \cite{nesbit_enhancing_2019}) angles, with 95\% front and and 80\% side overlap. Ground sampling distance (GSD) was approximately 3cm for all areas. Between four and six ground control points (GCPs) were placed in each area and precisely located using the GNSS receiver.
\\~\\
Ground Control Points (GCPs) were matched across images semi-automatically using Agisoft Metashape 1.8.1.13915 \mbox{\citep{agisoft_llc_agisoft_2022}}. An orthomosaic was generated for each area, also using Agisoft Metashape. These orthomosaics were cropped to remove edge effects. All data used to train deep learning-based ITC delineation were based on the orthomosaics rather than the original images. 

\subsubsection{Manual Delineation \& Training Data}
Orthomosaics for each area were split into 1024$\times$1024 tiles. This resolution was sufficiently large to preserve the visual context around each crown without downsampling, but small enough not to introduce excessive computational overhead. To create training data, all crowns visible in the orthoimagery were delineated manually, including those not covered by field plots, guided by geolocated trunk points where available. Manual delineation was performed  in the full-size orthomosaics and the resulting delineations split to match the tiled images. We did not delineate understory growth where it was visible in the drone imagery. A visual example can be seen in Figure~\ref{fig:resultpicslabels}. We make this tiled data available in the ubiquitous COCO format \citep{lin_microsoft_2015}. A summary table describing our data can be seen in Table~\ref{tab:data}. Note that these statistics are derived from the manually delineated polygons, rather than the inventory data, as the inventory data were not available for all trees, only those in the subplots within each orthomosaic. We did not perform any hyperparameter tuning, and left hyperparameters as specified by the original author for each component of our methodology (\cite{he_deep_2015}; \cite{he_mask_2017}; \cite{akyon_slicing_2022}). We therefore split our data into training and test sets only using nine-fold cross validation, geographically, with each of the nine areas corresponding to the orthomosaic generated by one of the nine non-overlapping drone flights (see Section~\ref{sec:droneimagery}). In this approach, each fold uses one specific area as the test set while the remaining areas were used for training. This methodology means that the individual predicted footprints, which we used to evaluate segmentation performance and estimate dieback, were not generated by a single, universal model. Instead, each prediction was produced by a model trained on eight areas and tested on the remaining area (from which the tile to predict on is drawn). The average performance of our segmentation approach and the dieback estimation comparison was thus calculated by aggregating the results from each of these nine distinct models, on the area that each model was not trained on, providing a performance metric that reflects performance across all areas.

\begin{table}[h]
\tabcolsep=0pt%
\TBL{\caption{Summary information for our dataset of tree crowns in the nine areas covered by drone flights in Almorox, Spain. Crown statistics are derived from manual delineations, rather than inventory, as ground sample plots only covered a small area of each orthomosaic. \label{tab:data}}}
{\begin{fntable}
\begin{tabular*}{\textwidth}{@{\extracolsep{\fill}}lcccccc@{}}\toprule%
    \TCH{Area No.} & \TCH{Area (m$^2$)} & \TCH{Dimensions (Px)}   & \TCH{No. Crowns} & \TCH{Mean Crown Area (m$^2$)} & \TCH{No. Tiles} \\
    \midrule
    1 & 17,942 & 5491$\times$8169 & 87 & 51.4 & 49  \\    
    2 & 28,516 & 6837$\times$10427 & 298 & 28.7 & 72  \\     
    3 & 28,269 & 8263$\times$8553 & 366 & 22.1 & 77  \\     
    4 & 19,225 & 5835$\times$8237 & 326 & 27.1 & 56  \\       
    5 & 23,288 & 7489$\times$7774 & 141 & 50.5 & 60  \\     
    6 & 16,663 & 6308$\times$6604 & 202 & 25.1 & 45  \\      
    7 & 39,736 & 8781$\times$11313 & 391 & 20.1 & 108 \\      
    8 & 31,001 & 8599$\times$9013 & 175 & 50.9 & 77  \\    
    9 & 39,020 & 9945$\times$9809 & 177 & 84.1 & 104 \\      
    \midrule
    \textbf{Total/Avg.} & 243661 & N/A & 2163 & 40.0 & 648    \\  
    \botrule
\end{tabular*}%
\end{fntable}}
\end{table}

\subsection{Methods}    
\subsubsection{ITC Delineation via Instance Segmentation}
To delineate individual trees from tiled images, we used the ubiquitous Mask R-CNN framework \citep{he_mask_2017} with a ResNet-101 FPN backbone \citep{he_deep_2015}. The backbone was pretrained on ImageNet \citep{deng_imagenet_2009}, and both random resizing and flipping were used for augmentation. We henceforth refer to crown footprints predicted on individual tiled images as `tiled predictions'.
\\~\\
Tiled predictions were recombined using Slicing Aided Hyper Inference (SAHI; \cite{akyon_slicing_2022}). We did not change the hyperparameters, used tiles with a relative overlap of 0.2 at inference, and post-processed predictions using greedy Non-Maximum Merging (NMM) with an Intersection-Over-Smaller (IOS) match threshold of 0.5. During ITC delineation (instance segmentation), each predicted crown is given a confidence score, corresponding to how likely it is that it contains a crown. A minimum confidence threshold of 0.3 was imposed on predictions before merging, as in the original SAHI implementation \citep{akyon_slicing_2022}. We report segmentation performance using mean Average Precision (mAP; see \cite{beitzel_encylcopedia_2009}) at two stages - on the tiled dataset, averaged across all tiles within each area, and after the tiled predictions are recombined using SAHI, for each of the nine orthomosaics.

\subsubsection{Vegetation Index\mbox{-}Based Dieback Estimates at the Tree Level}
We used an iterative approach to match ITCs to GNSS-measured trunk locations for trees with ground-truth GPS field locations (positional error < 1 m) and expert dieback assessment (percentage crown defoliation) in the full-size orthomosaics. To do this, the Euclidean distance was calculated between the centroid of each ITC and all ground truth trunk locations. The pair with the smallest distance was taken to be a match, and both ITC and ground truth (field) location for that pair were not considered as part of further matches. This step was repeated until no ground truth locations remained. We opted to discard ITC-trunk pairs where the distance from the centroid of the predicted crown to the ground-truth trunk location was greater than the square root of the crown area, to avoid including automatic dieback estimates that did not reliably correspond to the supposedly matching in-situ estimates. See Appendix~\ref{app:crownmatching} for pseudocode. We repeated this analysis twice - once using the ITCs predicted using the deep learning-based ITC delineation model, and once using the manually delineated ITCs (which were also used to train the segmentation model).
\\~\\
For the matched crown-trunk pairs, defoliation was estimated via Green Chromatic Coordinate (GCC), a common metric that has been used to track phenology successfully in a range of ecosystems (\cite{richardson_tracking_2018}), and a similar metric to the vegetation indices used to track tree health in other works (\mbox{\cite{reid_using_2016}}; \mbox{\cite{sandric_trees_2022}}). Green Chromatic Coordinate is defined below in Equation~\ref{eq:gcc}. 
\begin{equation}
    \label{eq:gcc}
    \text{GCC} = \frac{G}{R+G+B}
\end{equation}
Here, R, G and B refer to the total red, green and blue pixel values respectively, summed across the region of interest. Individual tree crowns were matched to ground-surveyed trunk locations (see Appendix~\ref{app:crownmatching}), and drone-derived GCC values were then correlated with field-based percentage of defoliation, using GCC values derived from both the manually labelled and automatically segmented crowns.

\section{Results}\label{sec:results}
\begin{figure}[]
\centering 

\begin{subfigure}{5cm}
\includegraphics[width=\linewidth]{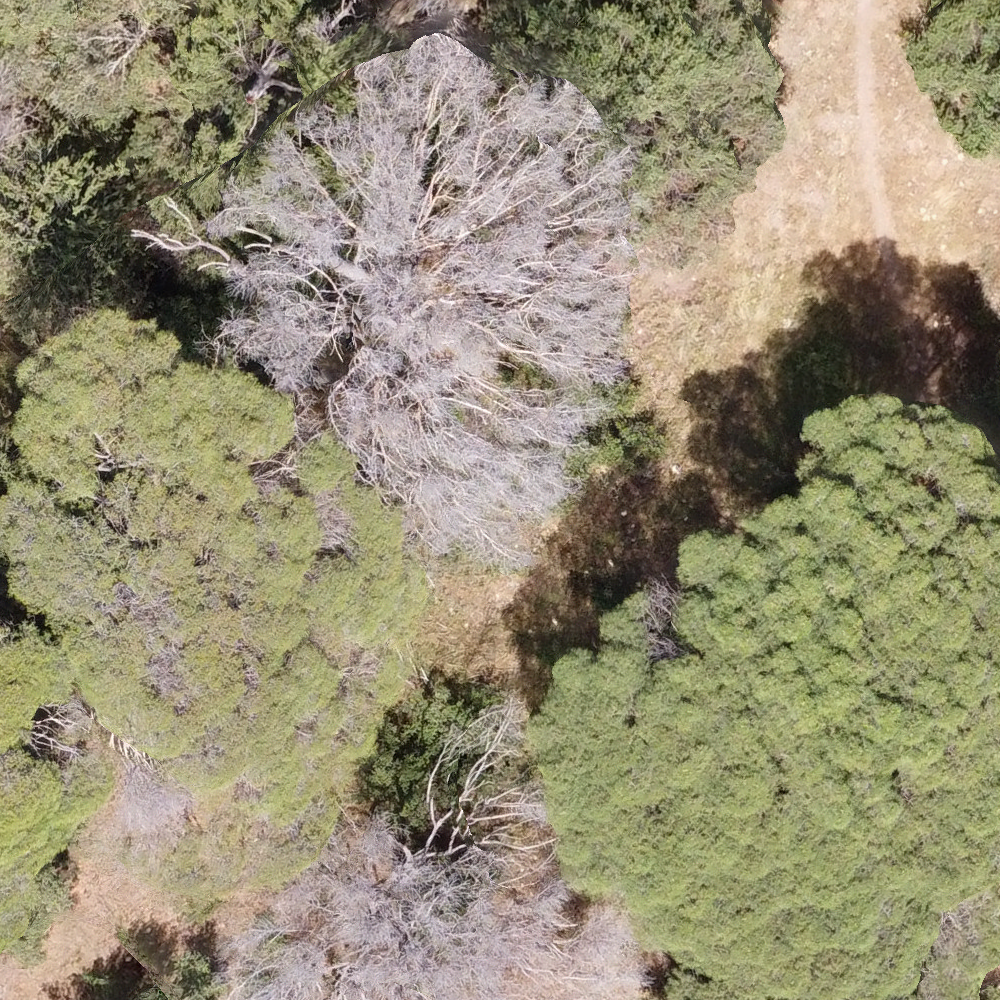}
\caption{}
\label{fig:resultpicsblank}
\end{subfigure}
\begin{subfigure}{5cm}
\includegraphics[width=\linewidth]{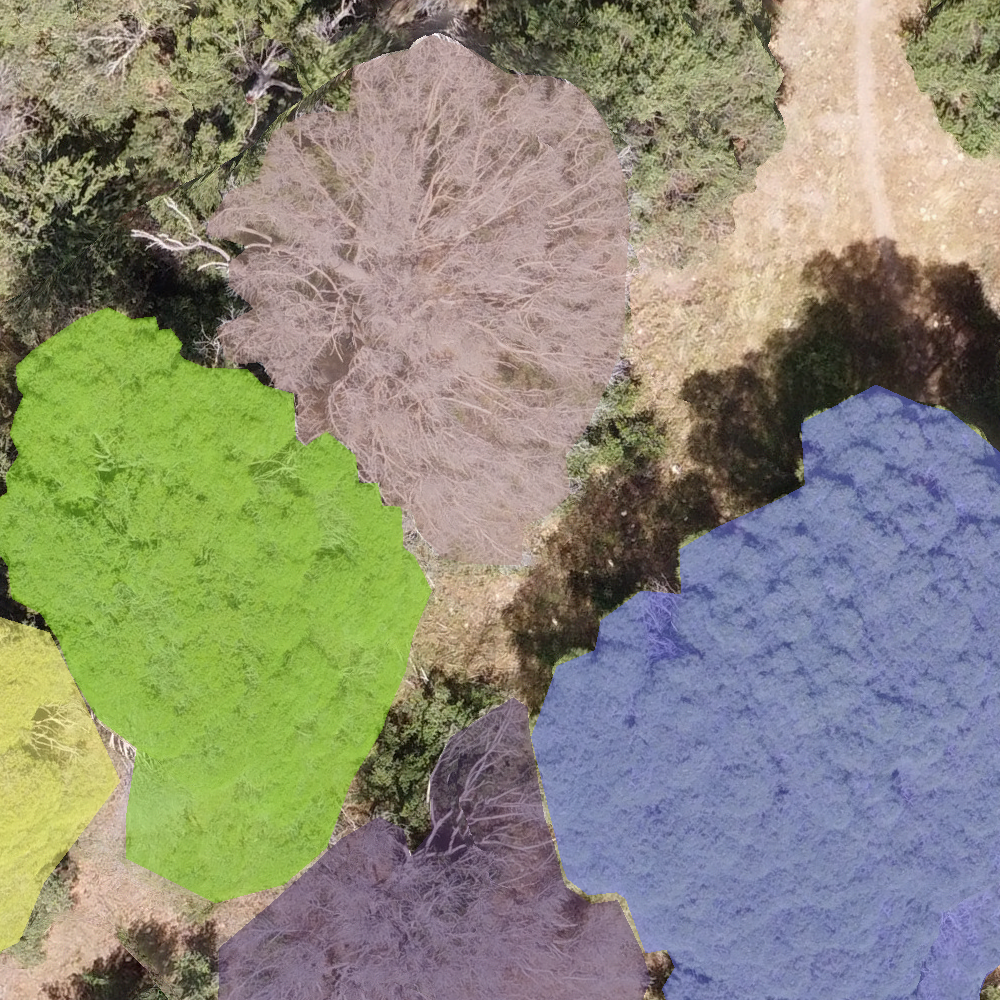}
\caption{}
\label{fig:resultpicslabels}
\end{subfigure}
\begin{subfigure}{5cm}
\includegraphics[width=\linewidth]{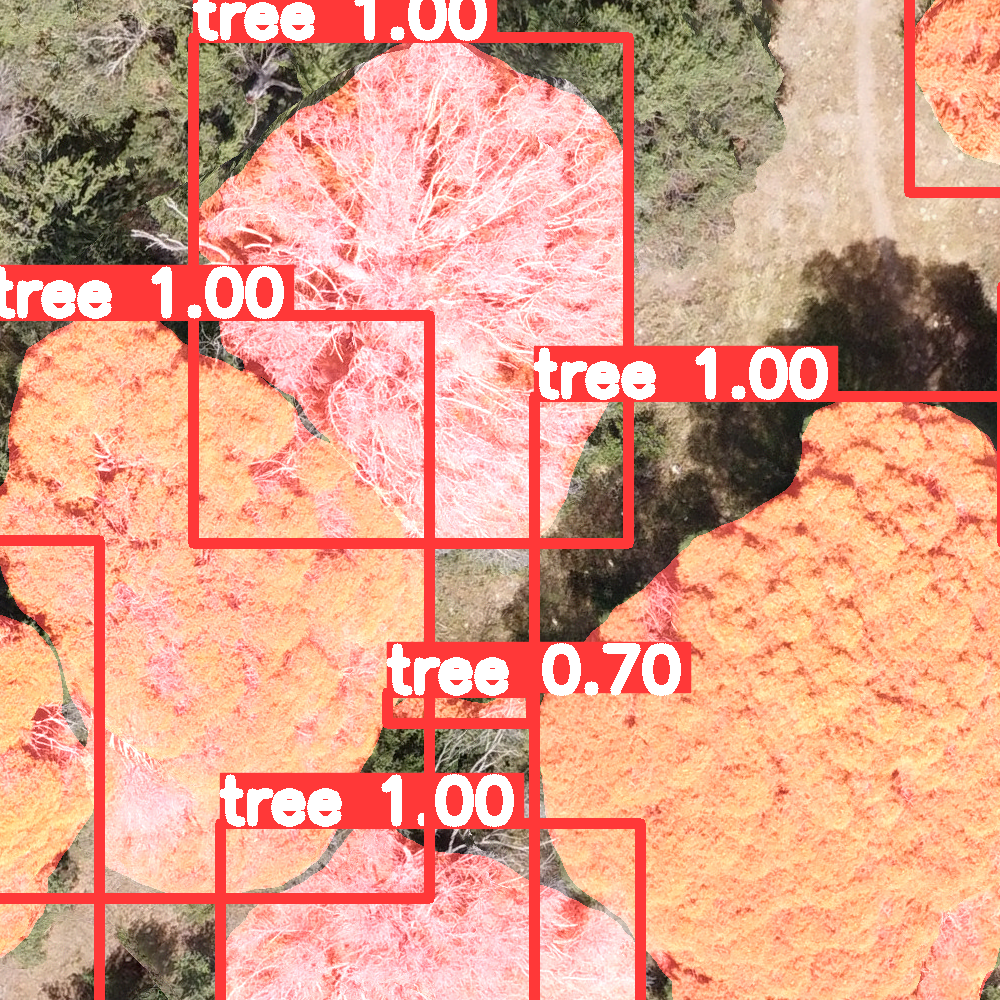}
\caption{}
\label{fig:resultpicspredictions}
\end{subfigure}

\caption{\textbf{(a)} Unlabelled, \textbf{(b)} manual, and \textbf{(c)} automatically predicted crown polygons for both healthy crowns (left, bottom right) and crowns exhibiting dieback (top centre, bottom centre). Numbers next to the class name 'tree' denote confidence score corresponding to each prediction. Images in this figure span approximately 30m.}
\label{fig:resultpics}
\end{figure}

\begin{figure}[] 
\centering 

\begin{subfigure}{\textwidth}
\includegraphics[width=14.5cm]{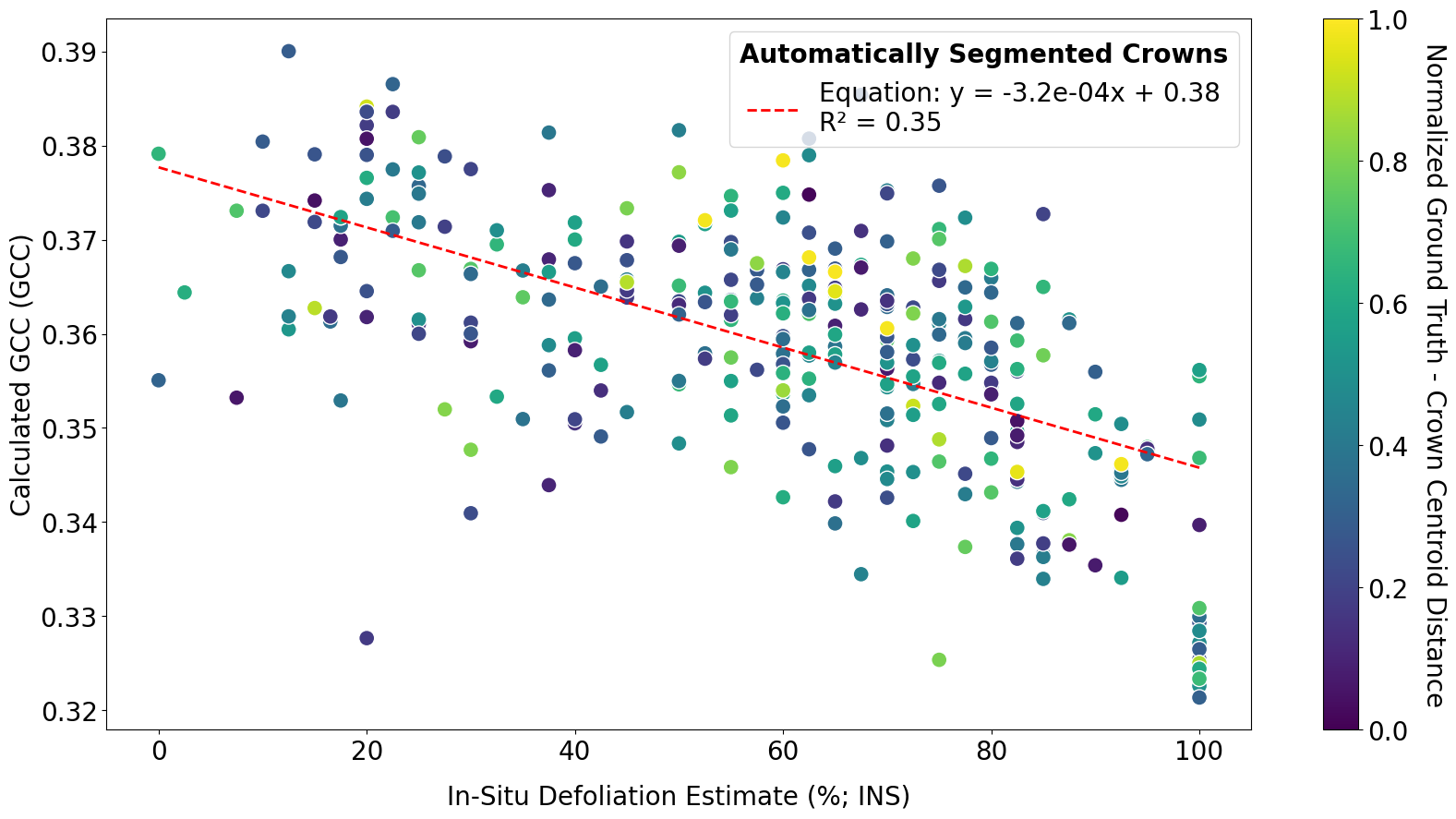}
\caption{Plot of Estimated Green Chromatic Coordinate (GCC) vs. field-based defoliation (percentage of defoliation), using \textbf{automatically segmented crowns}. Linear model derived using Ordinary Least Squares (OLS), $p<4\times10^{-32}$. Since we used nine-fold cross validation, each data point is produced by one of nine models, each of which has not been influenced by the ground-truth trunk location to which it is matched (as it corresponds to a tree in the area comprising the test set).}
\label{fig:resultcorrelationpred}
\end{subfigure}
\begin{subfigure}{\textwidth}
\includegraphics[width=14.5cm]{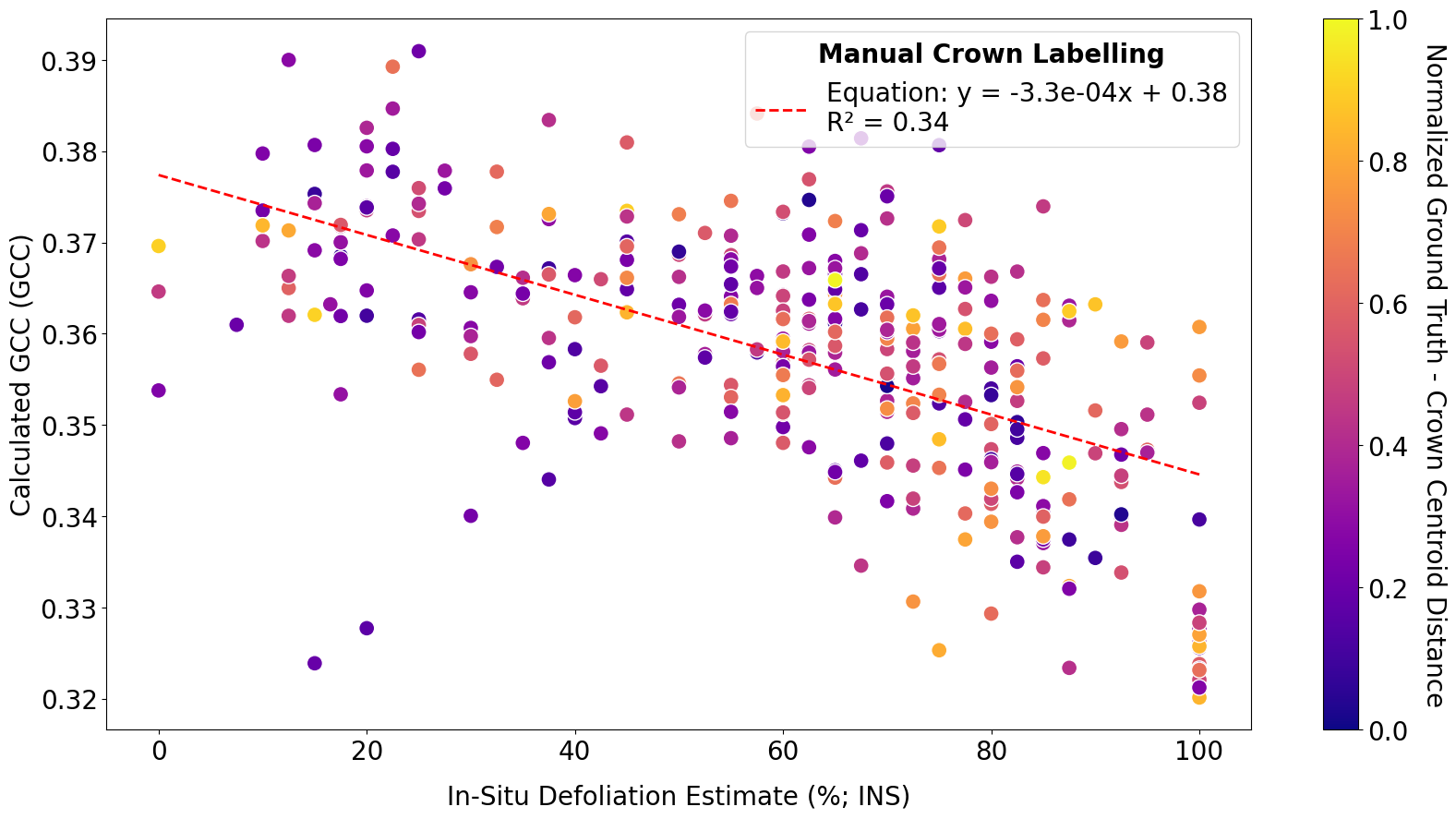}
\caption{Plot of Estimated Green Chromatic Coordinate (GCC) vs.field-based defoliation (percentage of defoliation, using \textbf{manually segmented crowns}. Linear model derived using Ordinary Least Squares (OLS), $p<2\times10^{-33}$.}
\label{fig:resultcorrelationgt}
\end{subfigure}

\caption{Dispersion plots of Estimated Green Chromatic Coordinate (GCC) vs. field-based defoliation at the individual tree level.}
\label{fig:resultcorrelation}
\end{figure}

\begin{figure}[] 
\centering 
\includegraphics[width=14.5cm]{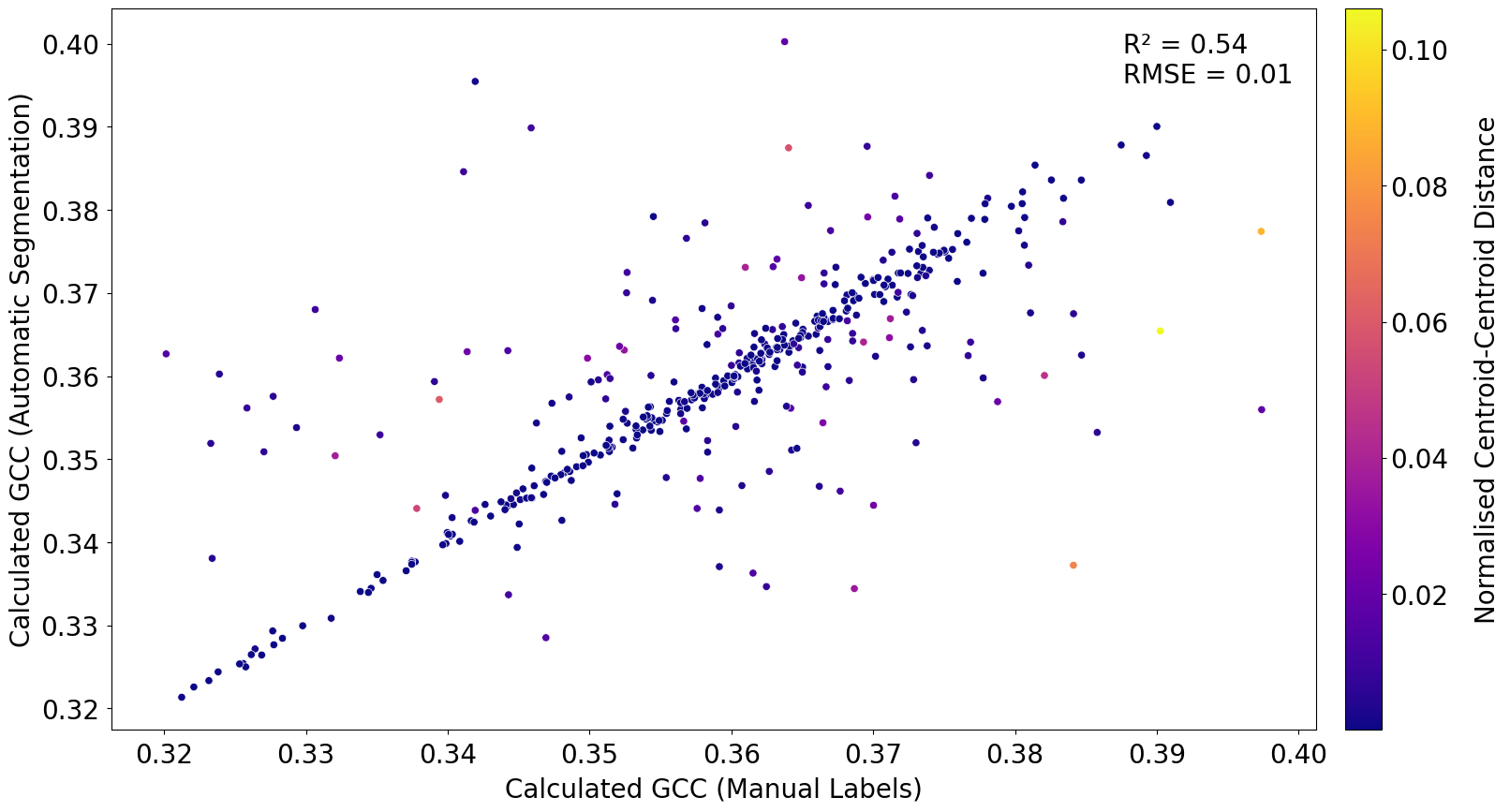}
\caption{Correlation between Green Chromatic Coordinate (GCC) estimated using manually labelled vs. automatically segmented crowns. Crowns are matched according to the corresponding inventory trunk location from the crown matching algorithm. The RMSE between the two GCC estimates was found to be 0.01, with $p<3\times10^{-72}$.}
\label{fig:gt_auto_correlation}
\end{figure}

Our results show that automatic crown delineation was possible in variably packed canopy with low species variation (Figure~\ref{fig:resultpics}). We present both per-tile averages and results for full-size orthomosaic inference using SAHI, per-area in Table~\ref{tab:segresults}. Visual results can be seen in Figure~\ref{fig:resultpics}. An average mAP of $0.519$ was achieved on the tiled data, with a minimum of $0.473$, a maximum of $0.602$ and a standard deviation of $0.037$. When recombined for the full orthomosaic using SAHI, mean mAP was reduced to $0.433$, with a minimum of $0.377$, maximum of $0.544$ and a standard deviation of $0.065$. We note that performance varied somewhat from area to area. We also note that recombining the tiled predictions increased variation in performance, and decreased mAP for all nine areas.
\\\\
The correlation between drone and field-based defoliation estimation was strong (Figure~\ref{fig:resultcorrelation}). We correlated ground estimates of defoliation with GCC estimates derived from crowns segmented in aerial imagery, using both crowns segmented using deep learning (Figure~\ref{fig:resultcorrelationpred}) and crowns segmented by hand (Figure~\ref{fig:resultcorrelationgt}). For the automatically segmented crowns, we found the correlation between calculated GCC and field-based defoliation estimates was significant ($p<4\times10^{-32}$, R$^{2}=0.35$). When the analysis was repeated using the manually segmented (ground truth) crown polygons (Figure~\ref{fig:resultcorrelationgt}, the correlation was significant with equivalent performance ($p<2\times10^{-33}$, R$^{2}=0.34$).
\\\\
There was little change in the strength of correlation between field-based defoliation estimation and aerial GCC-based estimates, when calculated with the automatically extracted crowns (Figure~\ref{fig:resultcorrelationpred}) versus the manually labelled crowns (Figure~\ref{fig:resultcorrelationgt}). The GCC estimates for each case were matched according to the corresponding ground truth trunk location obtained for each polygon via Algorithm~\ref{alg:gcd}, and showed strong correlation against each other in Figure~\ref{fig:gt_auto_correlation} (R$^{2}=0.54$, RMSE$=0.01$, $p<3\times10^{-72}$).

\begin{table}[]
\tabcolsep=0pt%
\TBL{\caption{Summary results (crown segmentation mAP) as mean plus/minus standard deviation for ITC delineation of our dataset from Almorox, Spain using Nine-fold cross validation. Mean average precision is reported at IoU thresholds of 0.5, 0.75 and averaged between 0.5 and 0.95 using intervals of 0.05, as in \cite{he_mask_2017}. \label{tab:segresults}}}
{\begin{fntable}
\begin{tabular*}{\textwidth}{@{\extracolsep{\fill}}lcccccc@{}}\toprule%
                 & \multicolumn{3}{@{}c@{}}{\TCH{Tiled (Mask R-CNN)}} & \multicolumn{3}{@{}c@{}}{\TCH{Orthomosaic (Mask R-CNN + SAHI)}} \\
\cmidrule{2-4}\cmidrule{5-7}%
\TCH{Area No.} & \TCH{mAP} & \TCH{mAP$_{0.5}$} & \TCH{mAP$_{0.75}$} & \TCH{mAP} & \TCH{mAP$_{0.5}$} & \TCH{mAP$_{0.75}$} \\\midrule
1 & 0.602 & 0.792 & 0.681 & 0.490 & 0.740 & 0.518 \\    
2 & 0.507 & 0.715 & 0.564 & 0.424 & 0.630 & 0.458 \\     
3 & 0.558 & 0.777 & 0.641 & 0.544 & 0.779 & 0.610 \\     
4 & 0.496 & 0.749 & 0.547 & 0.439 & 0.696 & 0.483 \\       
5 & 0.515 & 0.763 & 0.563 & 0.315 & 0.544 & 0.343 \\     
6 & 0.522 & 0.756 & 0.576 & 0.471 & 0.695 & 0.537 \\      
7 & 0.486 & 0.702 & 0.545 & 0.460 & 0.653 & 0.501 \\      
8 & 0.473 & 0.661 & 0.525 & 0.377 & 0.580 & 0.379 \\    
9 & 0.512 & 0.763 & 0.563 & 0.377 & 0.661 & 0.325 \\      
\midrule
\textbf{Mean $\pm$ std.} & 0.519$\,\pm\,$0.037 & 0.742$\,\pm\,$0.039 & 0.578$\,\pm\,$0.047 & 0.433$\,\pm\,$0.065 & 0.664$\,\pm\,$0.070 & 0.462$\,\pm\,$0.090 \\  
\botrule
\end{tabular*}%
\end{fntable}}
\end{table}

\newpage
\section{Discussion}\label{sec:discussion}

We found that, when evaluated on both the tiled and full-orthomosaic crown segmentation data, Mask R-CNN produced strong results (mAP = $0.519\pm0.037$), comparable or higher to those on machine learning benchmark data (mAP = $0.371$ from \cite{lin_microsoft_2015}). Although results on different data are not directly comparable, these relatively similar mAP scores indicate that ITC delineation in monospecific canopy is no less achievable than segmenting everyday objects from curated imagery - although we stress that the absolute accuracy of the manually annotated training labels is not known. The lack of need for additional technical development underscores the potential of models such as Mask R-CNN for non-expert use, proving their applicability to real-world conservation. Whilst segmentation performance could likely be improved with a more sophisticated or larger model and better postprocessing, these may not be the limiting factors to using such models in a practical context. Performance decreased in almost all cases when recombining tiled to full-size orthomosaic prediction using SAHI.
\\\\ 
Direct comparison to previous work performing ITC delineation is difficult due to variation in metrics used to assess performance, differences in canopy structural complexity between ecosystems, and the inclusion or omission of ground-based data to verify labels. Ground-based data serve as a more definitive source of accuracy, and the lack of such data in some studies (\cite{yang_detecting_2022}; \cite{ball_accurate_2023}) can potentially lead to misleading performance metrics or less reliable interpretations of model effectiveness. Although mean average precision (mAP) and similar derivative metrics are widely accepted for measuring instance segmentation performance on large-scale benchmark datasets such as COCO \citep{lin_microsoft_2015}, these metrics are not commonly reported in similar works - with authors commonly opting to report performance using metrics such as F1-score \citep{ball_accurate_2023} or precision \citep{weinstein_individual_2019}. The interpretability of these metrics is beneficial, but these are point metrics relying on a single selection of minimum confidence and Intersection-over-Union (IoU) thresholds - which can be adjusted arbitrarily to maximise the target metric \citep{maxwell_accuracy_2021b}. A higher or lower value of F1-score, precision or recall may not represent a better or worse model for a user, even when trained on the same data. mAP-based metrics are a better reflection of model performance, and higher values indicate that a model is likely to be robust to different manual selections of confidence and IoU thresholds (\cite{maxwell_accuracy_2021a}; \cite{maxwell_accuracy_2021b}). Given the complex nature of forests, including variation in canopy structural complexity, a model's performance across different ecosystems and conditions relies on its ability to handle observed variability.
\\\\
We observed a significant negative correlation between field-based defoliation estimates and drone-derived GCC in our site for both automatic (R$^2=0.35$; Figure~\ref{fig:resultcorrelationpred}) and manual segmentation (R$^2=0.34$; Figure~\ref{fig:resultcorrelationgt}), with a high correlation between estimation using the manual labels vs the automatically segmented crowns (Figure~\ref{fig:gt_auto_correlation}). Our findings suggest that even a basic deep learning approach performs as effectively as manual annotations in producing ITC footprints for dieback estimation. Investing in more advanced segmentation may not offer significant improvements for estimating dieback or, perhaps, other crown metrics such as leaf area index (LAI) - although further verification would be required to confirm this. Manual labels are, however, difficult to verify as being accurate, and it may be the case that training the model based on crowns labelled by humans from above introduces operator bias (\cite{bai_finding_2018}; \cite{geva_are_2019}). Verifiable ground-truth labels could derived, for example, from additional instruments such as Terrestrial Laser Scanners (TLS), but such data was not available for this site. Whether it is possible to match the accuracy of TLS-based segmentation from RGB imagery alone is unknown at this time.
\\\\
The variance in calculated GCC for a given field-based defoliation percentage may be derived from physical reasons related to field-based estimates. Dieback is not expected to be uniform within each crown \citep{denman_chapter_2022} - so metrics based on per-pixel colour averages, without regard to spatial patterns, are unlikely to capture the degree of dieback for each crown with perfect accuracy. This effect is exaggerated by the particular patterns of drought-induced dieback, which typically begins at the extremities of the crown \citep{denman_chapter_2022} - resulting in substantially different observations when viewed from above, as in the aerial images used here, and from below, as would be seen in-situ. Although other, similar work, uses different vegetation indices \mbox{\citep{sandric_trees_2022}}, we suggest that the variance in results is unlikely to diminish significantly while still using vegetation indices based on simple colour space transformations. Extracting dieback estimates using the deep learning model directly - akin to classification in the context of the original Mask R-CNN model \citep{he_mask_2017}, but modified to perform regression - may improve performance as these estimates could leverage spatial patterns in addition to colour information. We opted not to use this approach to avoid the need to produce dieback labels for trees visible in orthoimagery that were not already covered by existing field-based estimates.
\\\\
Our results show that, for data taken with the same instruments, vegetation index-based estimates correlate significantly with expert field-based estimates. However, care should be taken when making direct comparisons between estimates for individual trees. For example - comparing dieback estimates for the same tree from imagery taken at different times, or with different instruments, may generate confusion as measured colour intensities will change due to lighting or camera differences. We suggest that colour calibrating imagery may improve this, although we did not perform colour calibration here as all images were taken with the same instruments and did not make direct comparisons between individual data points obtained under different conditions.
\\\\
We successfully automated canopy defoliation measurement in a single-species canopy, comprising \textit{P. pinea}, the only species displaying dieback in our ecosystem. Minor modifications are likely to be required for our approach - based on vegetation indices applied to crown footprints extracted using deep learning-based instance segmentation - to be applicable to data of multiple species. Since canopy reflectance may vary due to differences in leaf spectra and crown structure corresponding to species \citep{ollinger_sources_2011}, we would not recommend directly comparing vegetation index-based estimates between two species. The approach of \mbox{\citet{sandric_trees_2022}} indicates that using the deep learning model to classify crown footprints by species prior to any health monitoring is promising, although this increases the labelling burden on practitioners. It is possible that an increase in open ITC delineation data may help to relieve this problem, by removing the need for site-specific datasets, or by reducing volume requirements on labels via pretraining on such open data. Alternatively, estimating dieback based on spatial patterns rather than simple colour-space based estimation may prove to be more comparable, as dieback is not expected to be uniform within each crown \citep{denman_chapter_2022}. In the case of our data, the canopy comprises \textit{P. pinea} almost exclusively - which is also the only species displaying dieback in this ecosystem - meaning we do not suffer from such confusion in this work.
\\\\
Our approach has three important advantages that decrease financial and manual cost to application. Firstly, an increase in processing speed, secondly, the use of a low cost, widely available drone platform, and thirdly, the reduced reliance on expert analysis in the field. The time savings, for example, of using a large-scale aerial approach such as this rather than a ground method, are significant. Unlike a typical field campaign, including manual inventory and perhaps the use of other instruments such as laser scanners, drone flights for a site can be conducted in hours to days rather than weeks to months. With a trained model, postprocessing these data to extract downstream metrics requires only a few hours (for a dataset of this size on consumer hardware). Short, drone-based campaigns also have smaller requirements for material - with the DJI Mavic Mini used here being available for only a few hundred euros, and being light enough to fly with minimal legal requirements. Combined with the reduced number of person-hours required, the cost of conducting and processing data from a drone-based survey is minimal compared with a traditional field campaign.
\\\\
Although we used significant manual input at two points during postprocessing, this could be avoided in practice as more open data becomes available \citep{lines_ai_2022}. Firstly, manual Ground Control Points (GCPs) were used to stitch together the orthorectified images. We found orthophotos stitched together without the use of GCPs were of comparable visual quality, although we did not compare the use of these images downstream to avoid the need to repeat the time consuming labelling process. It is possible that the orthomosaics may be georeferenced less accurately without the use of GCPs, although this could be addressed by mounting a more accurate GPS receiver to the drone, and such RTK systems are increasingly becoming available on consumer drone systems. Secondly, the manual labelling process was time consuming, and required several weeks to complete by a single user. Although we used ground truth labels to verify our results, they may not be required for practitioners to use similar approaches, particularly with community efforts to provide ground-truth training data for similar applications \citep{puliti_forinstance_2023}. Large foundation models, for example, can produce reasonable segmentation in a wide range of unseen contexts when pretrained on vast quantities of general visual data \citep{kirillov_segment_2023}, although can be difficult to employ using consumer hardware. Some of these approaches rely on context-specific prompting to identify rough areas of interest \citep{kirillov_segment_2023}. As accuracy requirements for prompting are less strict than for full segmentation, methods for generating these prompts developed on open forest data may be sufficiently accurate on unseen ecosystems. Alternatively, smaller models such as the one used here could be pretrained on public ITC delineation data, and used without retraining. Both of these solutions, however, require large-scale data of tree crowns from a diverse set of ecosystems.
\\\\
In addition to speed and cost savings, the adoption of automated, digital methods also benefits accuracy \citep{mu_characterization_2018} and reduces dataset sampling bias. Typically, an inventory-based approach can gather only simple structural measurements such as trunk diameter or height, which could be used to produce crude estimates of difficult to measure quantities such as crown area. A data-driven approach, using drone-based remote sensing and deep learning, allows for measurement of quantities such as crown area far more accurately than hand measurement, but without the laborious time requirements of other highly accurate instruments such as TLS scanners. The comprehensive nature of automated surveying also reduces selection bias - for example, vegetation in areas that are difficult to reach by foot, such as on very steep terrain, may differ physiologically from plants found on flat ground. A drone-based survey can include a large number of stems from such areas, whereas an inventory-based survey may only include stems in areas that can be accessed directly.
\\\\
With these combined improvements in speed, cost, coverage, and accuracy, profound implications for forest management in deteriorating ecosystems are evident. Real-time detection of tree dieback is essential, as declining trees become hotspots for primary pests and pathogens to thrive and multiply. Additionally, such trees are more susceptible to secondary pest attacks \citep{balla_threat_2021}. Regular updates on tree health are vital for guiding management strategies, such as thinning stands to lessen resource competition among surviving trees - crucial under extreme climate conditions - or implementing sanitation cuts to eliminate trees harbouring pest or pathogen populations \citep{roberts_effect_2020}.
\\\\
For this paradigm shift towards automation to be applied in practice, two requirements remain. Firstly, an increase in extensive, diverse open datasets is vital for both training and validation, minimising the labelling burden on end-users \citep{lines_ai_2022}. Secondly, open and easy-to-use implementations of these tools, applied to common forest management data formats, are needed to minimise barriers to deployment for non-expert users. Some promising initiatives already exist (\cite{ball_accurate_2023}; \cite{community_environmental_2023}). With these two requirements fulfilled, the shift to automation in forest management and conservation could be expedited. Focus should therefore be given to fostering a culture of data sharing, and to the development of accessible platforms to deploy these methods. Towards these goals, we make both code and data available for this work, in both common remote sensing and machine learning-readable formats.

\section{Conclusions}\label{sec:conclusions}
In this work, we made three key contributions to the state-of-the-art in aerial tree health assessment. Firstly, we demonstrated the feasibility of deep learning-based Individual Tree Crown (ITC) delineation in structurally complex monospecific natural canopies, where reduced leaf spectra variation between individual trees may result in relatively lower variation in canopy reflectance compared to mixed canopies. Although verifying crown segmentation precisely is challenging, we corroborated our segmentation method using ground truth GPS trunk locations to assess its accuracy. We highlight the need for further verification in future work, potentially through ground-based tools such as Terrestrial Laser Scanning (TLS).
\\\\
Secondly, our work found that detecting crown dieback using vegetation indices derived from deep learning-based crown footprints is feasible, as suggested in \citet{sandric_trees_2022}. For the first time, we validated this type of tree health assessment with field-based expert observations, and found that there was strong correlation between our automatic estimates and field-based estimates.
\\\\
Lastly, we give evidence that our method of tree health assessment is not overly sensitive to segmentation accuracy. Repeating the analysis with ground truth segmentation labels in place of model predictions did not significantly improve the correlation with field-based assessment. This finding suggests developing more precise segmentation methods for tree health assessment may not be necessary, although more verification is required when using more complex or spatially explicit measurements downstream.

\newpage

\begin{appendix}
\section{Residual Plots}

\begin{figure}[H] 
\centering 
\includegraphics[width=14.5cm]{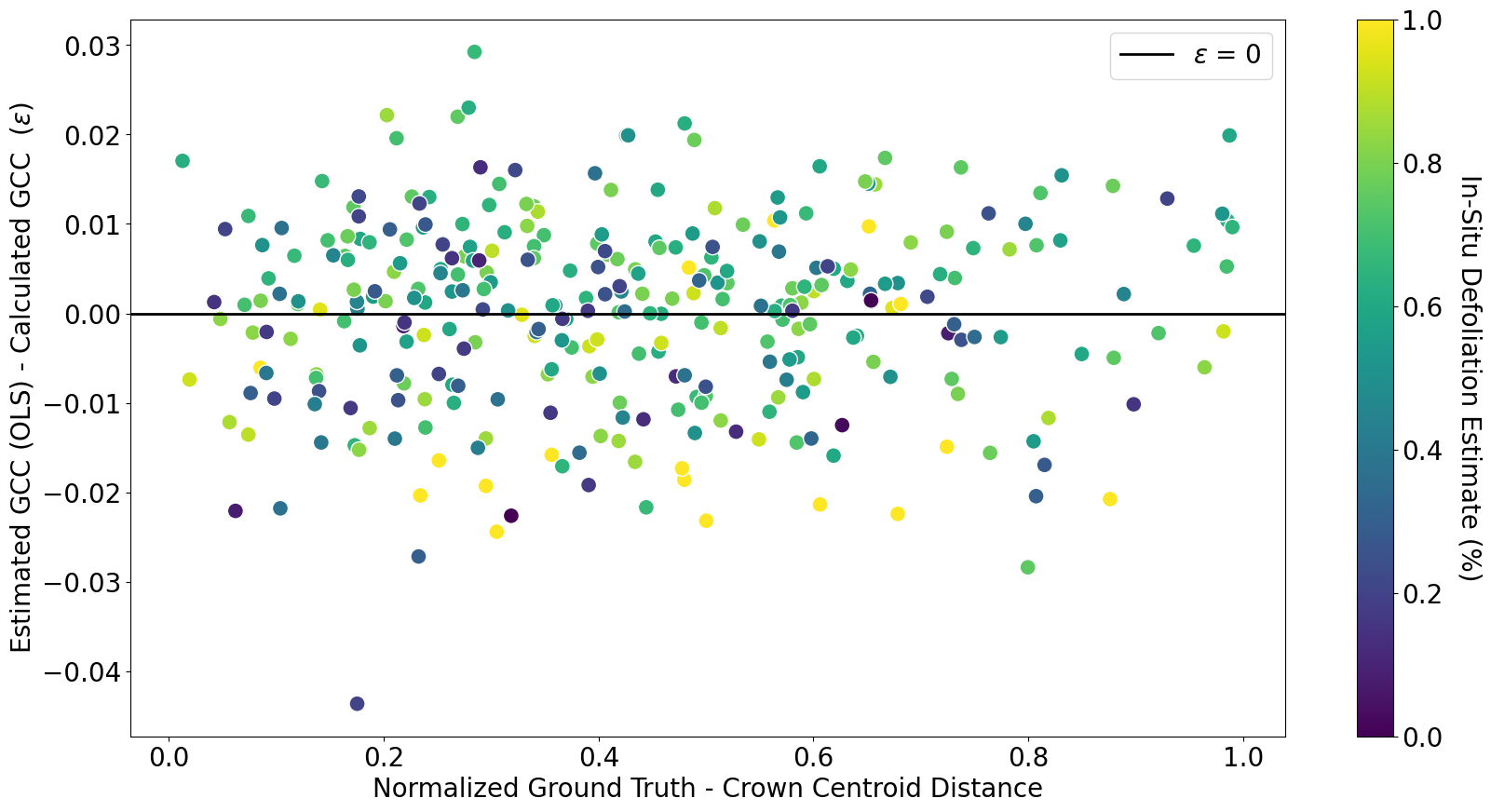}
\caption{Plot of GCC residuals from OLS vs. Distance from inventory trunk location to matched polygon.}
\label{fig:resultresidual}
\end{figure}

A plot of the residuals corresponding to the OLS of Figure~\ref{fig:resultcorrelation}, as a function of centroid-trunk match distance, can be seen in Figure~\ref{fig:resultresidual}. The magnitude of the residuals does not seem to increase with match distance for distances less than $1$, as per Algorithm~\ref{alg:gcd}. We suggest, for this reason, that the matching procedure outlined in Algorithm~\ref{alg:gcd} is not the source of any variation in calculated GCC. This is despite the observation that \textit{P. pinea} trunks often display curvature where the crown centre does not coincide with the trunk position. The linear model overestimates GCC for larger in-situ defoliation estimates. There is no reason a priori to expect GCC to be a linear function of the in-situ visual defoliation estimates. 
 
\newpage

\section{Crown Matching}\label{app:crownmatching}

We outline the iterative approach used to match individual tree crowns with ground-surveyed trunk locations in Algorithm~\ref{alg:gcd}. Refer to \url{https://numpy.org/doc/} for further details regarding specific functions \citep{harris2020array}.

\begin{algorithm}[H]
\caption{NumPy-like pseudocode for crown matching}
\label{alg:gcd}
\begin{algorithmic}
\State \textcolor{blue}{\# pred\_polys - list of predicted polygons}
\State \textcolor{blue}{\# n\_trunks - number of ground-truth trunk locations}
\State \textcolor{blue}{\# n\_poly - number of ITCs}
\\
\State $\textit{dist} \gets \texttt{np.empty}(\textit{n\_trunks}, \textit{n\_poly})$ 
\\
\While{\textit{dist.shape}[0] > 0} 
\State $\textit{i, j} \gets \texttt{np.where}(
\textit{dist} == \texttt{np.min}(\textit{dist}))$
\State $\textit{gcc} \gets \texttt{calculate\_gcc}(\textit{pred\_polys}\text{[}\textit{j}\text{]})$
\State $\textit{dist} \gets \texttt{np.delete}(\textit{dist}, \,\textit{i}, \,\textit{axis\text{=}0})$
\State $\textit{dist} \gets \texttt{np.delete}(\textit{dist}, \,\textit{j}, \,\textit{axis\text{=}1})$
\State $\textit{pred\_polys}.\texttt{pop}(\textit{j})$
\EndWhile

\end{algorithmic}
\end{algorithm}

\end{appendix}

\begin{Backmatter}

\paragraph{Abbreviations}
RGB (Red, Green, Blue); LiDAR (Light Detection And Ranging); ITC (Individual Tree Crown); NEON (National Ecological Observation Network); R-CNN (Region-based Convolutional Neural Network); CHM (Canopy Height Model); nDSM (Normalised Digital Surface Model); LSTM (Long Short-Term Memory); RTK (Real-Time Kinematic); GNSS (Global Navigation Satellite System); GSD (Ground Sampling Distance); GCP (Ground Control Point); COCO (Common Objects in COntext); SAHI (Slicing Aided Hyper Inference); NMM (Non-Maximum Merging); IOS (Intersection-Over-Smaller); mAP (Mean Average Precision); GCC (Green Chromatic Coordinate); ExG (Excess Green Index); RMSE (Root-Mean-Square Error); TLS (Terrestrial Laser Scanning); OLS (Ordinary Least Squares); IoU (Intersection-Over-Union); LAI (Leaf Area Index)

\paragraph{Funding Statement}
M. J. A. was supported by the UKRI Centre for Doctoral Training in Application of Artificial Intelligence to the study of Environmental Risks [EP/S022961/1]. E. R. L. and S. W. D. G. were funded by a UKRI Future Leaders Fellowship awarded to E. R. L. [MR/T019832/1]. PRB was supported by the Community of Madrid Region under the framework of the multi-year Agreement with the University of Alcalá (Stimulus to Excellence for Permanent University Professors, EPU-INV/2020/010) . PRB and ERL are supported by the Science and Innovation Ministry (subproject LARGE and REMOTE, N\textdegree~PID2021-123675OB-C41 and PID2021-123675OB-C42).

\paragraph{Competing Interests}
The author(s) declare none.

\paragraph{Data Availability Statement}
Our code is available at \href{https://github.com/mataln/dlncs-dieback}{github.com/mataln/dlncs-dieback}. A permanent record is available at \href{https://doi.org/10.5281/zenodo.10657079}{doi.org/10.5281/zenodo.10657079}. Our data is available at \href{https://doi.org/10.5281/zenodo.10646992}{doi.org/10.5281/zenodo.10646992}.

\paragraph{Ethical Standards}
The research meets all ethical guidelines, including adherence to the legal requirements of the study country.

\paragraph{Author Contributions}
All authors contributed to the conceptualization of the work; M. J. Allen and E. R. Lines designed the methodology; E. R. Lines, D. Moreno-Fernández and Ruiz-Benito collected and pre-processed the data; M. J. Allen processed the data and wrote the relevant software. M. J. Allen analysed the data and led the writing of the manuscript. All authors contributed critically to the drafts and gave final approval for publication.

\end{Backmatter}

\printbibliography

\end{document}